\documentclass{llncs}
\usepackage{llncsdoc}
\usepackage{times}
\usepackage{algorithm}
\usepackage{algorithmic}
\usepackage{hyperref}
\usepackage{amsfonts,mathrsfs}
\usepackage{amsmath,amssymb,graphicx,textcomp,enumerate,bbm,latexsym,xcolor}
\usepackage{graphicx,graphics,wrapfig,epstopdf,subfig}

\newcommand{\beq}{\begin{equation}}
\newcommand{\eeq}{\end{equation}}
\newcommand{\beqs}{\begin{eqnarray}}
\newcommand{\eeqs}{\end{eqnarray}}
\newcommand{\barr}{\begin{array}}
	\newcommand{\earr}{\end{array}}

\newcommand{\Amat}{{\bf A}}
\newcommand{\Bmat}{{\bf B}}

\newcommand{\Dmat}{{\bf D}}
\newcommand{\Emat}{{\bf E}}
\newcommand{\Fmat}{{\bf F}}

\newcommand{\Hmat}{{\bf H}}
\newcommand{\Imat}{{\bf I}}

\newcommand{\Smat}{{\bf S}}
\newcommand{\Tmat}{{\bf T}}
\newcommand{\Umat}{{\bf U}}
\newcommand{\Vmat}{{\bf V}}
\newcommand{\Wmat}{{\bf W}}
\newcommand{\Xmat}{{\bf X}}

\newcommand{\Zmat}{{\bf Z}}

\newcommand{\cv}{{\boldsymbol c}}
\newcommand{\dv}{{\boldsymbol d}}

\newcommand{\hv}{{\boldsymbol h}}

\newcommand{\sv}{{\boldsymbol s}}

\newcommand{\uv}{{\boldsymbol u}}
\newcommand{\vv}{{\boldsymbol v}}

\newcommand{\xv}{{\boldsymbol x}}
\newcommand{\yv}{{\boldsymbol y}}

\newcommand{\zv}{{\boldsymbol z}}

\newcommand{\alphav}{{\boldsymbol \alpha}}

\newcommand{\gammav}{{\boldsymbol \gamma}}

\begin{document}

\title{Compressive Sensing via Convolutional Factor Analysis}

\titlerunning{Compressive Sensing via Convolutional Factor Analysis}




%\author{Yizhe Zhang \and Changyou Chen \and Ricardo Henao \and Lawrence Carin\\
%Duke university, Durham NC \\
%yz196@duke.edu}	
%
%\authorrunning{Zhang, Chen, Henao, Carin}
%
%\institute{Department of Electrical \& Computer Engineering, Duke University\\
%\texttt{\{yz196,cc448,rhenao,lcarin\}@duke.edu}\\
%}

\author{Xin Yuan\inst{1}, Yunchen Pu\inst{2}  \and Lawrence Carin\inst{2}
}

\institute{Bell Labs, Nokia\\
	\texttt{xyuan@bell-labs.com}
	\and
Department of Electrical \& Computer Engineering, Duke University \\
}

\maketitle
\begin{abstract} 
We solve the compressive sensing problem via convolutional factor analysis, where the convolutional dictionaries are learned {\em in situ} from the compressed measurements. 
An alternating direction method of multipliers (ADMM) paradigm for compressive sensing inversion based on convolutional  factor analysis is developed.
The proposed algorithm provides reconstructed images as well as features, which can be directly used for recognition ($e.g.$, classification) tasks.
When a deep (multilayer) model is constructed, a stochastic unpooling process is employed to build a generative model. During reconstruction and testing, we project the upper layer dictionary to the data level and only a single layer deconvolution is required. 
We demonstrate that using $\sim30\%$ (relative to pixel numbers) compressed measurements, the proposed model achieves the classification accuracy comparable to the original data on MNIST.
We also observe that when the compressed measurements are very limited ($e.g.$, $<10\%$), the upper layer 
dictionary can provide better reconstruction results than the bottom layer. 
\end{abstract}
\section{Introduction}
The compressive sensing (CS) problem~\cite{cs_Candes06,Candes08L1,cs_Donoho06} can be formulated as:
\begin{eqnarray}
\min \frac{1}{2}\|\yv - \Amat\xv\|^2_2 + \lambda \|\cv\|_\star,\qquad
{\text{s.t.}}~~ \xv = \Bmat \cv, \label{eq:L1}
\end{eqnarray}
where $\Amat \in {\mathbb R}^{M\times N}$ is the sensing matrix and usually $M\ll N$. $\xv$ is the desired signal, $\zv$ denotes the coefficients which are sparse ($\|\cdot\|_\star = \|\cdot\|_1$)~\cite{cs06DonohoL1eqL0} or low rank ($\|\cdot\|_\star$ symbolizes the nuclear norm)~\cite{Dong14TIP}, and given $\cv$, we can recover $\xv$ via $\Bmat$.
This $\Bmat$ can be {\em known a priori}, ({\em e.g.}, a wavelet or DCT basis) or learned from the measurement $\yv$ during reconstruction. $\lambda$ is a parameter to balance the two terms in (\ref{eq:L1}).

There has been over a decade research on CS, both on theory and applications. Various algorithms have been proposed~\cite{Bioucas-Dias2007TwIST,daubechies2010iteratively,Figueiredo07GPSR} for CS inversion. On the other hand, deep learning methods, especially the convolutional~\cite{LeCun89CN} and deconvolutional networks~\cite{Zeiler10CVPR}, have achieved excellent recognition results on benchmark datasets~\cite{GoogleNet,Simonyan15ICLR}. The convolutional networks, used in a supervised manner, usually aim to extract good features to achieve high classification performance.
By contrast, the deconvolutional networks~\cite{Zeiler10CVPR,Pu2016AISTATS,Chen13deepCFA}, used in an {\em unsupervised} manner, aim to reconstruct the input signals ({\em e.g.}, minimize the reconstruction error to the input images), as well as extracting features.
Therefore, it is applicable to use this deconvolutional network to solve the CS problem in~(\ref{eq:L1}).

Using {\em in situ} learned dictionaries to solve the CS problem is not new, while most existing algorithms learn dictionaries on small patches~\cite{MairalICML09,Yu11SPT}. 
Compressive sensing, however, usually imposes compression on the entire image~\cite{Duarte08SPM,Huang13ICIP}. Similarly, the convolutional factor analysis~\cite{Chen13deepCFA,Pu2016AISTATS} models learn dictionaries on entire images, too.
Thereby, it is feasible and appropriate to leverage this convolutional factor analysis technique to reconstruct desired signals $\xv$ from compressed measurements $\yv$. 
Regarding the regularizer, in convolutional factor analysis, the coefficients (a.k.a., features) are usually imposed to be sparse and therefore the $\ell_1$-norm is utilized in~(\ref{eq:L1}).

This paper makes the following contributions:
$i$) A new convolutional factor analysis algorithm based on {\em compressed measurements} is developed using the alternating direction method of multipliers (ADMM) paradigm~\cite{Boyd11ADMM}.
$ii$) As the features are obtained during reconstruction, our algorithm provides features simultaneously with reconstruction results. Therefore, joint classification and reconstruction is straightforward.
$iii$) Via using the stochastic ``unpooling" approach~\cite{Pu2016AISTATS}, we constitute a deep generative model without losing information.
$iv$) We project the upper layer dictionary down to the data level and following this, only a single layer deconvolution is required for both image reconstruction and classification. 
We demonstrate that using $\sim30\%$ (relative to pixel numbers) compressed measurements, we can achieve the classification performance comparable to the original data on MNIST.
As an additional observation, we notice that compressive sensing with random sensing matrix can improve the classification results, which is in agreement with the recent theory developed in~\cite{Huang15ICASSP,Huang15NIPS}.

In the following of this paper, we first derive a new convolutional factor analysis (CFA) algorithm in Section~\ref{Sec:CFA_admm}. This regime is extended to the CS case in Section~\ref{Sec:CS_cfa}, where the CFA is performed on the compressed measurements directly, thus {\em in situ} convolutional dictionary learning. 
A deep model using stochastic unpooling~\cite{Pu2016AISTATS} is constructed in Section~\ref{Sec:deep}.  
Joint modeling of reconstruction and classification is briefly introduced in Section~\ref{Sec:Joint_model}. Experimental results are presented in Section~\ref{Sec:result}.
Though a lot of implementation details will be unveiled in the paper, hereby we emphasize that since in the CS problem, the target is to recover the original signal, we impose the feature size larger than the image size such that the desired image is always within the {\em valid}  region of the convolutional operation. This is different from the feature extraction in deconvolutional learning~\cite{Pu2016AISTATS} and convolutional networks~\cite{LeCun89CN}, where the feature lies in the valid region of the convolution.  
\section{Convolutional Factor Analysis via ADMM \label{Sec:CFA_admm}}
Considering the image case investigated in CS, let ${\Xmat_n}\in{\mathbb R}^{N_x \times N_y \times N_c}$ denote the three-dimensional (3D) image, which can be a gray-scale image ($N_c = 1$), an RGB image ($N_c = 3$), or a hyperspectral image ($N_c$ denoting the spectral channels).
Under the convolutional factor model, we jointly consider $N$ images, and for $n^{th}$ image
\begin{equation}
{\textstyle  \Xmat_n = \sum_{k=1}^K \Dmat_k * \Smat_{k,n} + \Emat_n},
\end{equation}
where $\Dmat_k \in {\mathbb R}^{n_x \times n_y \times N_c}$ is the convolutional dictionary (kernels or filters), $\Smat_{k,n} \in {\mathbb R}^{(N_x +n_x-1) \times (N_y +n_y-1) \times N_c}$ denotes the coefficients (features) and $\Emat_n$ signifies the residual or noise. 
The two-dimensional convolution `$*$' is performed on each slice ($n_c = 1,\dots, N_c$) of $\Dmat_k$ and  $\Smat_{k,n}$.
Note the spatial size of $\Smat_{k,n}$ is $(N_x +n_x-1) \times (N_y +n_y-1)$ such that the image $\Xmat_n$ will be of `valid' size after convolution.
As mentioned before, this is different from the convolutional neural networks, which impose the features to be of valid size.
We develop the CFA algorithm based on ADMM below, which is different from~\cite{Zeiler10CVPR}. Note that the dictionaries $\{\Dmat_k\}_{k=1}^K$ are shared across $N$ images, while the features $\{\Smat_{n,k}\}_{k,n = 1}^{N,K}$ vary for each image.

Sparsity is imposed on $\Smat$ to solve the problem. Without considering compressive sensing, the problem during training can be modeled as
\begin{equation} \label{eq:conv_p1}
\min ~\frac{1}{2} {\textstyle \sum_{n=1}^N \|\Xmat_n -\sum_{k=1}^K \Dmat_k * \Smat_{k,n}\|_2^2 + \lambda \|\Smat\|_1},
\end{equation}
where $\|\Smat\|_1 = \sum_{n,k} \|\Smat_{k,n}\|_1$.
Equation (\ref{eq:conv_p1}) results in the following objective function
\begin{equation}
{\cal L}(\Dmat, \Smat, \lambda) = \frac{1}{2} \sum_{n=1}^N \|\Xmat_n -\sum_{k=1}^K \Dmat_k * \Smat_{k,n}\|_2^2 +\lambda \sum_{n,k} \|\Smat_{k,n}\|_1.
\end{equation}

In order to simplify the problem, we introduce an auxiliary variable $\Zmat$, and the problem in (\ref{eq:conv_p1}) can be formulated as
\begin{align}
\min ~~{\textstyle \frac{1}{2} \sum_{n=1}^N \|\Xmat_n -\sum_{k=1}^K \Dmat_k * \Smat_{k,n}\|_2^2 + \lambda \|\Zmat\|_1, }
\qquad{\text{s.t.}} ~~\Zmat = \Smat.\label{eqn:Pconv1}
\end{align}
Consider the Lagrange multiplier $\{{\Vmat}, \eta\}$ and denote
$\sv = {\rm vec}(\Smat), \zv = {\rm vec}(\Zmat), \vv = {\rm vec}(\Vmat)$. This leads to another objective function
\begin{align}\label{eqn:Pconv2}
&{\cal L}(\Dmat, \Smat, \Zmat,\Vmat, \lambda,\eta) = {\textstyle \frac{1}{2} \sum_{n=1}^N \|\Xmat_n -\sum_{k=1}^K \Dmat_k * \Smat_{k,n}\|_2^2} \nonumber \\
&\qquad {\textstyle +\lambda \|\Zmat\|_1 + \frac{\eta}{2} \|\Smat-\Zmat\|_2^2 + \vv^\top(\sv- \zv)}. 
\end{align}
Define $\Umat = (1/\eta)\Vmat$, and we have
\begin{align}%{eqn:Pconv2}
&{\textstyle{\cal L}(\Dmat, \Smat, \Zmat,\Umat, \lambda,\eta) 
	= \frac{1}{2} \sum_{n=1}^N \|\Xmat_n -\sum_{k=1}^K \Dmat_k * \Smat_{k,n}\|_2^2} \nonumber\\
&\qquad{\textstyle+\lambda \|\Zmat\|_1  + \frac{\eta}{2}\|\Smat - \Zmat +\Umat\|^2_2  -\frac{\eta}{2}\|\Umat\|_2^2 }.\label{eq:conv3}
\end{align}

ADMM cyclically solves (\ref{eq:conv3}) via the following sub problems:
\begin{align}
\Dmat^{t+1} &:= \arg \min_{\Dmat} (\frac{1}{2}\sum_{n=1}^N \|\Xmat_n -\sum_{k=1}^K \Dmat_k * \Smat_{k,n}\|_2^2) 
\label{eq:D_t+1}\\
\Smat^{t+1} &:= \arg \min_{\Smat} ({\textstyle\frac{1}{2}\sum_{n=1}^N \|\Xmat_n -\sum_{k=1}^K \Dmat_k * \Smat_{k,n}\|_2^2} {\textstyle+ \frac{\eta}{2}\|\Smat - \Zmat +\Umat\|^2_2}) 
\label{eq:S_t+1}\\
\Zmat^{t+1} &:= \arg \min_{\Zmat} {\textstyle(\lambda \|\Zmat\|_1 + \frac{\eta}{2} \|\Smat - \Zmat + \Umat\|_2^2) }
\label{eq:Z_t+1}\\
%\Umat^{t+1}&:= \arg \min_{\Umat} \frac{\eta}{2}\|\Smat - \Zmat +\Umat\|^2_2  -\frac{\eta}{2}\|\Umat\|_2^2
\Umat^{t+1}&:= \Umat^{t} + \eta (\Smat -\Zmat)
\label{eq:U_t+1}
\end{align} 
where $t$ denotes the iteration. Below we solve these subproblems one by one.
The complete ADMM-CFA algorithm is summarized in Algorithm~\ref{algo:ADMM-CFA}.
\begin{itemize}
	%\vspace{-2mm}
	\item[1)] Eq. (\ref{eq:D_t+1}) can be solved by gradient descent. Via the following definitions: a) $\xv_{n} = {\rm vec}(\Xmat_{n})$, b) $\dv_{k} = {\rm vec}(\Dmat_{k})$, c) $\Fmat_{k,n}$ being the 2D sparse convolution matrix that implements $\Dmat_k * \Smat_{k,n} = \Fmat_{k,n} \dv_{k}$, and d)
	%\begin{equation}
	${\xv}_n^{-k} = \xv_n -\sum_{k'=1, k'\neq k}^K \Fmat_{k',n} \dv_{k'}$,
	%\end{equation}
	taking derivative to $\dv_k$ in (\ref{eq:D_t+1}), we have
	\begin{align}
	&\frac{\partial (\frac{1}{2}\sum_{n=1}^N \|\xv_n -\sum_{k=1}^K\Fmat_{k,n} \dv_{k}\|_2^2) }{\partial \dv_{k}} = {\textstyle -\sum_{n=1}^N\Fmat_{k,n}^\top (\xv_n^{-k} -\Fmat_{k,n} \dv_{k})}.
	\end{align} 
	Therefore,
	\begin{equation} \label{eq:update_D}
	{\textstyle \dv_{k}^{t+1} = \dv_{k}^t + \beta \sum_{n=1}^N\Fmat_{k,n}^\top (\xv_n^{-k} -\Fmat_{k,n} \dv_{k})},
	\end{equation}
	where $\beta$ is the learning rate\footnote{Implementation details in MATLAB:
		\vspace{-2mm}
		\begin{align}
		\Fmat_{k,n} \dv_{k} &= \Dmat_k * \Smat_{k,n} = {\rm conv2}(\Smat_{k,n}, \Dmat_k, \text{`valid'}) \label{eq:Fmat_def}\\
		\Fmat_{k,n}^\top \xv_{n} &= {\rm conv2}({\text {rot90}}(\Smat_{k,n},2),\Xmat_{n},  \text{`valid'}) \\
		\Tmat_{k} \sv_{k,n} &= \Dmat_k * \Smat_{k,n} = {\rm conv2}(\Smat_{k,n}, \Dmat_k, \text{`valid'}) \label{eq:Tsv}\\
		\Tmat_{k}^\top \xv_{n} &= {\rm conv2}(\Xmat_{n},{\text {rot90}}(\Dmat_{k},2),  \text{`full'})
		\vspace{-3mm}
		\end{align} 
		We also found that using ``fft2(~)" in MATLAB is at least $4\times$ faster than ``conv2(~)" by providing almost the same results.}.
	\item[2)] Eq. (\ref{eq:S_t+1}) is a quadratic optimization problem and can be simplified via the following definitions: a) $\sv_{k,n} = {\rm vec} (\Smat_{k,n})$ and b) $\Tmat_{k,n}$ being the 2D sparse convolution matrix that implements $\Dmat_k * \Smat_{k,n} = \Tmat_{k} \sv_{k,n}$. Since the feature is unique for each image, given $\{\Xmat_n, \Dmat, \Zmat_n, \Umat_n\}$, we define
	\begin{align}
	{\cal C}_n &\stackrel{\rm def}{=} 
	{\textstyle\frac{1}{2} \|\xv_n -\sum_{k=1}^K \Tmat_{k} \sv_{k,n}\|_2^2}{\textstyle+ \frac{\eta}{2}\sum_{k}\|\sv_{k,n} - \zv_{k,n} +\uv_{k,n}\|^2_2}.
	\end{align}  
	Taking derivative to $\sv_{k,n}$, we have
	\begin{equation}
	\frac{\partial{\cal C}_n}{\partial\sv_{k,n}} = \eta (\sv_{k,n} - \zv_{k,n} +\uv_{k,n})- \Tmat_k^\top(\xv_n -\sum_{k=1}^K \Tmat_{k} \sv_{k,n}).
	\end{equation}
	Denote 
	${\xv}_n^{-k} = \xv_n -\sum_{k'=1, k'\neq k}^K \Tmat_{k'} \sv_{k',n}$,
	%\end{eqnarray}
	and set $	\frac{\partial{\cal C}_n}{\partial\sv_{k,n}} = (\Tmat_k^\top \Tmat_k + \eta \Imat) \sv_{k,n}  - (\Tmat_k^\top{\xv}_n^{-k} + \eta\zv_{k,n} -\eta\uv_{k,n}) = 0$.
	Given $\{\Tmat, \zv,\uv\}$, the optimal $\sv_{k,n}$ is the solution of the following linear system
	\begin{equation} \label{eq:update_S}
	(\Tmat_k^\top \Tmat_k + \eta \Imat) \sv_{k,n} = \Tmat_k^\top{\xv}_n^{-k} + \eta(\zv_{k,n} -\uv_{k,n}),
	\end{equation}
	which can be solved effectively
	using conjugate gradient (CG) algorithms~\cite{Shewchuk94CG}.
	\item[3)] Eq. (\ref{eq:Z_t+1}) can be solved via the shrinkage operator, {\em i.e.}, soft thresholding
	\begin{equation}
	{\textstyle  \Zmat^{t+1} = {\rm soft} (\Smat^t + \frac{\Umat^t}{\eta}, \gammav^t)},
	\end{equation}
	which can be performed element-wise, {\em i.e.}, for $i^{th}$ element,
	\begin{equation} \label{eq:update_Z}
	{\textstyle z^{t+1}_i = {\rm sign}(s^t_i + \frac{u^t_i}{\eta}) ~\max(|s^t_i + \frac{u^t_i}{\eta}|- \gamma^t_i,0)},
	\end{equation}
	and this threshold $\gammav^t$ can be updated in each iteration, thus {\em adaptively} soft thresholding. Note that $\lambda$ in (\ref{eq:Z_t+1}) can be normalized and thus observed in $\eta$.
	In our experiments, we set the number of the non-zero elements (thus sparsity) in $\Zmat$ to be the same in each iteration and update this $\gammav^t$ based on $(\Smat^t + {\Umat^t}/{\eta})$.
\end{itemize}

\begin{center}
	\begin{algorithm}[htbp]
		\caption{ADMM-CFA}
		\begin{algorithmic}[1]
			\REQUIRE Input images $\{\Xmat_n\}_{n=1}^N$, parameters $\{\beta,\eta\}$ 
			\STATE Initialize $\Dmat, \Smat, \Zmat, \Umat$.
			\FOR{$t=1$ {\bfseries to} MaxIter }
			\STATE Update $\Dmat$ by Eq. (\ref{eq:update_D}).
			\STATE Update $\Smat$ by Eq. (\ref{eq:update_S}).
			\STATE Update $\Zmat$ by shrinkage operator, Eq. (\ref{eq:update_Z}).
			\STATE Update $\Umat$ by Eq. (\ref{eq:U_t+1}).
			\ENDFOR
		\end{algorithmic}
		\label{algo:ADMM-CFA}
	\end{algorithm}
\end{center}

\section{Convolutional Factor Analysis Based Compressive Sensing~\label{Sec:CS_cfa}}
We now extend the CFA model to the CS scenario.
Consider $N$ images jointly with the same sensing matrix $\Amat$, (which can also be different for each image).
\begin{equation}
\yv_n = \Amat \xv_n,
\end{equation} 
where $\xv_n$ denotes the $n^{th}$ vectorized image.
Similar to (\ref{eq:L1}), using the CFA model, the problem can be formulated as
\begin{align}
\min &~~\frac{1}{2}\sum_{n=1}^N\|\yv_n - \Amat\xv_n\|^2_2 + \lambda \sum_{n=1}^N\sum_{k=1}^K\|\sv_{k,n}\|_1,\\
{\text{s.t.}} & ~~ {\textstyle \Xmat_n = \sum_{k=1}^K \Dmat_k * \Smat_{k,n}}. \label{eq:xTs}
\end{align}
Employing the definition in (\ref{eq:Tsv}), and using the vectorization forms, we  have
\begin{align}
\min &~~\frac{1}{2}\sum_{n=1}^N\|\yv_n - \Amat\xv_n\|^2_2 + \lambda \sum_{n=1}^N\sum_{k=1}^K\|\sv_{k,n}\|_1,\\
{\text{s.t.}} &~~ {\textstyle\xv_n = \sum_{k=1}^K \Tmat_k \sv_{k,n}} \label{eq:xTs_v}.
\end{align}
This leads to the following objective function
\begin{equation}\label{eq:CS_cfa}
{\cal L}(\Tmat, \Smat, \lambda) = \frac{1}{2} \sum_{n=1}^N \|\yv_n -\Amat \sum_{k=1}^K \Tmat_k \sv_{k,n}\|_2^2 +\lambda \sum_{n,k} \|\sv_{k,n}\|_1.
\end{equation}
We consider to solve (\ref{eq:CS_cfa}) in two ways:
%\begin{itemize}

a) The convolutional dictionary ($\{\Dmat_k\}_{k=1}^K$ or $\{\Tmat_k\}_{k=1}^K$) is pre-learned by training data. In this case, (\ref{eq:CS_cfa}) aims to solve 
\begin{equation}\label{eq:CS_cfa_preD}
\arg \min_{\{\sv\}}  \frac{1}{2} \sum_{n=1}^N \|\yv_n -\Amat \sum_{k=1}^K \Tmat_k \sv_{k,n}\|_2^2 +\lambda \sum_{n,k} \|\sv_{k,n}\|_1.
\end{equation}

b) The convolutional dictionary ($\{\Dmat_k\}_{k=1}^K$ or $\{\Tmat_k\}_{k=1}^K$) is unknown  {\em a priori} and will be learned {\em in situ} from the raw measurements $\{\yv_n\}_{n=1}^N$. In this case, (\ref{eq:CS_cfa}) aims to solve 
\begin{equation}\label{eq:CS_cfa_D}
\arg \min_{\{\Tmat,\sv\}}  \frac{1}{2} \sum_{n=1}^N \|\yv_n -\Amat \sum_{k=1}^K \Tmat_k \sv_{k,n}\|_2^2 +\lambda \sum_{n,k} \|\sv_{k,n}\|_1.
\end{equation} 
%\end{itemize}

\subsection{Compressive Sensing Inversion with Pre-Learned Convolutional Dictionary}
In this case, we aim to solve $\{\sv_{k,n}\}_{n,k= 1}^{N,K}$ given $\{\yv_n\}_{n=1}^N$ and $\{\Tmat_k\}_{k=1}^K$. We again employ the ADMM framework to solve this problem.
Recall (\ref{eqn:Pconv1}) and introducing the auxiliary variable $\{\zv_{n,k}\}_{n,k=1}^{N,K}$, we have the following problem
\begin{align}
\min \frac{1}{2} \sum_{n=1}^N \|\yv_n -\Amat \sum_{k=1}^K \Tmat_k \sv_{k,n}\|_2^2 + \lambda \sum_{k,n}\|\zv_{k,n}\|_1, 
\quad {\text{s.t.}} ~\sv_{k,n} = \zv_{k,n}, ~~\forall k,n.\label{eqn:Pconv_preD}
\end{align}
This results in the objective function
\begin{align} \label{eqn:Pconv_preD_admm}
&{\textstyle{\cal L}(\sv,\zv, \uv, \lambda,\eta) =\frac{1}{2} \sum_{n=1}^N \|\yv_n -\Amat\sum_{k=1}^K \Tmat_k \sv_{k,n}\|_2^2} \nonumber\\
&+\lambda \sum_{k,n}\|\zv_{k,n}\|_1 + \frac{\eta}{2}\sum_{k,n} \|\sv_{k,n}-\zv_{k,n} + \uv_{k,n}\|_2^2 +{\rm Const}. 
\end{align}
ADMM cyclically solves (\ref{eqn:Pconv_preD_admm}) via the following sub-problems:
{\small \begin{align}
	\sv^{t+1} &:= \arg \min_{\sv} {\textstyle(\frac{1}{2}\sum_{n=1}^N \|\yv_n -\Amat \sum_{k=1}^K \Tmat_k  \sv_{k,n}\|_2^2 }\nonumber\\
	&\qquad+ {\textstyle\frac{\eta}{2}\sum_{n,k}\|\sv_{k,n}-\zv_{k,n} + \uv_{k,n}\|^2_2) }
	\label{eq:S_t+1_preD}\\
	\zv^{t+1} &:= \arg \min_{\zv} (\lambda \sum_{k,n}\|\zv_{k,n}\|_1 \nonumber\\
	&\qquad+ {\textstyle\frac{\eta}{2} \sum_{k,n}\|\sv_{k,n}-\zv_{k,n} + \uv_{k,n}\|_2^2)} 
	\label{eq:Z_t+1_preD}\\
	%\Umat^{t+1}&:= \arg \min_{\Umat} \frac{\eta}{2}\|\Smat - \Zmat +\Umat\|^2_2  -\frac{\eta}{2}\|\Umat\|_2^2
	\uv^{t+1}&:= \uv^{t} + \eta (\sv -\zv)
	\label{eq:U_t+1_preD}
	\end{align} }
Note that the updates of $\zv,\uv$ in (\ref{eq:Z_t+1_preD}-\ref{eq:U_t+1_preD}) are the same as in (\ref{eq:Z_t+1})-(\ref{eq:U_t+1}).
For (\ref{eq:S_t+1_preD}), it is different from  (\ref{eq:D_t+1}). 
Following similar derivations, via defining 
\begin{align}
{\yv}_n^{-k} &\stackrel{\rm def}{=} {\textstyle\yv_n -\Amat\sum_{k'=1, k'\neq k}^K \Tmat_{k'} \sv_{k',n}},\nonumber
\end{align}
given $\{\Tmat, \zv,\uv,\Amat\}$, the optimal $\sv_{k,n}$ is the solution of the following linear system
\begin{equation} \label{eq:update_S_preD}
(\Tmat_k^\top \Amat^\top \Amat\Tmat_k + \eta \Imat) \sv_{k,n} = \Tmat_k^\top\Amat^\top{\yv}_n^{-k} + \eta(\zv_{k,n} -\uv_{k,n}).
\end{equation}
Again, this can be solved effectively
via CG algorithms.

\begin{center}
	\begin{algorithm}[htbp]
		\caption{CS-CFA}
		\begin{algorithmic}[1]
			\REQUIRE Input measurements $\{\yv_n\}_{n=1}^N$, sensing matrix $\Amat$, parameters $\{\beta,\eta\}$ 
			\STATE Initialize $\Dmat, \Smat, \Zmat, \Umat$.
			\FOR{$t=1$ {\bfseries to} MaxIter }
			\STATE Update $\Dmat$ by Eq. (\ref{eq:CS_dt+1}).
			\STATE Update $\Smat$ by Eq. (\ref{eq:update_S_preD}).
			\STATE Update $\Zmat$ by shrinkage operator, Eq. (\ref{eq:update_Z}).
			\STATE Update $\Umat$ by Eq. (\ref{eq:U_t+1_preD}).
			\ENDFOR
		\end{algorithmic}
		\label{algo:CS_CFA}
	\end{algorithm}
\end{center}
\subsection{Compressive Sensing Inversion by Learning Convolutional Dictionary from Measurements}
When the pre-learned dictionary is not available, we need to learn the dictionary from the raw measurements $\{\yv_n\}_{n=1}^N$.
The problem in Eq. (\ref{eq:CS_cfa_D}) results in the following objective function
\begin{align} \label{eqn:Pconv_D_admm}
&{\textstyle {\cal L}(\sv,\zv, \uv, \Tmat, \lambda,\eta)=\frac{1}{2} \sum_{n=1}^N \|\yv_n -\Amat\sum_{k=1}^K \Tmat_k \sv_{k,n}\|_2^2} \nonumber\\
&+\lambda \sum_{k,n}\|\zv_{k,n}\|_1 + \frac{\eta}{2}\sum_{k,n} \|\sv_{k,n}-\zv_{k,n} + \uv_{k,n}\|_2^2 +{\rm Const}. \nonumber
\end{align}

Recalling the definition in (\ref{eq:Fmat_def}), in addition to the subproblems described in (\ref{eq:S_t+1_preD})-(\ref{eq:U_t+1_preD}), we also need to update $\{\Dmat_k\}_{k=1}^K$, 
\begin{equation}
\dv^{t+1} := \arg \min_{\dv} (\frac{1}{2}\sum_{n=1}^N \|\yv_n -\Amat \sum_{k=1}^K \Fmat_{k,n}\dv_k \|_2^2).
\label{eq:d_t+1}
\end{equation}
Similar to (\ref{eq:D_t+1}), this can be solved by the gradient descent:
\begin{equation}
{\textstyle \dv_{k}^{t+1} = \dv_{k}^t + \beta \sum_{n=1}^N\Fmat_{k,n}^\top \Amat^\top(\yv_n^{-k} -\Amat\Fmat_{k,n} \dv_{k})}, \label{eq:CS_dt+1}
\end{equation}
where
${\yv}_n^{-k} = \yv_n -\Amat\sum_{k'=1, k'\neq k}^K \Fmat_{k',n} \dv_{k'}$.
The complete algorithm of compressive sensing convolutional factor analysis (CS-CFA) is summarized in Algorithm~\ref{algo:CS_CFA}.

\section{Going Deep: Multi-Layer Convolutional Factor Analysis \label{Sec:deep}}
The algorithm developed above, without compressive sensing, is similar to the deconvolutional networks~\cite{Zeiler10CVPR} (a single layer model), but different in the formulation as we are using the ADMM framework.
The Bayesian version of this deep deconvolutional networks, has recently been proposed by~\cite{Pu2016AISTATS} using a novel stochastic ``unpooling" approach, which is significant different from~\cite{Zeiler10CVPR}.
The key of this ``unpooling" is to impose that inside each pooling block (from the $\ell^{th}$ layer to $(\ell+1)^{th}$ layer), there is at most one nonzero element (only one nonzero element or all zeros), which will be pooled to the $(\ell+1)^{th}$ layer as input. Therefore, this ``unpooling" process does not lose any information from the bottom layer (touching the data) to the top layer.
Also, based on this, we can reconstruct images from upper layer factor analysis as all the information inside the measurements is kept in the deep model. We develop this {\em deep CS-CFA} below.

Consider an $L$-layer model and we demonstrate this deep model via two consecutive layers, {\em i.e.}, $\ell^{th}$ layer and $(\ell +1)^{th}$ layer.
\begin{align}
\Xmat^{\ell}_n = {\textstyle \sum_{k_{\ell}=1}^{K_{\ell}} \Dmat^{\ell}_{k_{\ell}} * \Smat^{\ell}_{k_{\ell},n}} + \Emat_n^{\ell},\qquad
\Smat^{\ell}_{n} = {{\mathsf {unpool}}} (\Xmat^{\ell+1}_n).
\end{align}
Note that for the first layer, because of the CS, we have
\begin{equation}
\yv_n = \Amat \xv^{1}_n + \Emat_n^1,
\end{equation}
and this $\{\yv_n\}_{n=1}^N$ is the only data we have and will be used to train the model (for both image reconstruction and feature extraction).
Keep performing CFA on the $(\ell+1)^{th}$ layer. We have
\begin{equation}
{\textstyle \Xmat^{\ell+1}_n = \sum_{k_{\ell+1}=1}^{K_{\ell+1}} \Dmat^{\ell+1}_{k_{\ell+1}} * \Smat^{\ell+1}_{k_{\ell+1},n} + \Emat_n^{\ell+1}}.
\end{equation}
To explicitly describe this ``unpool" process, we introduce the index variable $\Wmat_n^{\ell}\in\{0,1\}$, which has the same size of $\Smat_{n}^{\ell}$.
Recall the dimensions
%\begin{eqnarray}
$\Xmat^{\ell}_n \in {\mathbb R}^{N_x^{\ell} \times N_y^{\ell} \times K_{\ell-1}},
\Dmat^{\ell} \in  {\mathbb R}^{n_x^{\ell} \times n_y^{\ell} \times K_{\ell}},
\Smat^{\ell}_n \in  {\mathbb R}^{(N^{\ell}_x + n^{\ell}_x -1) \times (N^{\ell}_y + n^{\ell}_y -1) \times K_{\ell}},
\Wmat_n^{\ell}\in\{0,1\}^{(N^{\ell}_x + n^{\ell}_x -1) \times (N^{\ell}_y + n^{\ell}_y -1) \times K_{\ell}}$.
%\end{eqnarray}
Consider non-overlapping pooling blocks from $\Smat^{\ell}_n$ to $\Xmat^{\ell+1}_n$ and the block size is $p_x^{\ell} \times p_y^{\ell}$. Assuming integral division, 
\begin{equation}
{\textstyle \Xmat^{\ell+1}_n \in  {\mathbb R}^{(N^{\ell}_x + n^{\ell}_x -1)/p_x^{\ell} \times (N^{\ell}_y + n^{\ell}_y -1)/p_y^{\ell} \times K_{\ell}} }.
\end{equation}
For $k_{\ell}^{th}$ slice of $\Xmat^{\ell+1}_n$, $\Xmat^{\ell+1, k_{\ell}}_n \in {\mathbb R}^{(N^{\ell}_x + n^{\ell}_x -1)/p_x^{\ell} \times (N^{\ell}_y + n^{\ell}_y -1)/p_y^{\ell}}$, we define the $(i,j)^{th}$ element as $X^{\ell+1, k_{\ell}}_{n,i,j}, i = 1,\dots, (N^{\ell}_x + n^{\ell}_x -1)/p_x^{\ell}; j = 1,\dots, (N^{\ell}_y + n^{\ell}_y -1)/p_y^{\ell}$.
This $X^{\ell+1, k_{\ell}}_{n,i,j}$ corresponds to $(p_x^{\ell} \times p_y^{\ell})$ elements in $\Smat^{\ell,k_{\ell}}_n$, and  $(p_x^{\ell} \times p_y^{\ell})$ elements in $\Wmat^{\ell,k_{\ell}}_n$. 
We denote them as
\begin{eqnarray}
{\textstyle \{S^{\ell,k_{\ell}}_{n,i,j,r} \}_{r=1}^{p_x^{\ell} \times p_y^{\ell}}} \in {\mathbb R}^{p_x^{\ell} p_y^{\ell}},\qquad
{\textstyle \{W^{\ell,k_{\ell}}_{n,i,j,r} \}_{r=1}^{p_x^{\ell} \times p_y^{\ell}} }\in \{0,1\}^{p_x^{\ell} p_y^{\ell}}.
\end{eqnarray}
As mentioned above, each block will have {\em at most} one non-zero element, and we thus impose
\begin{eqnarray}
W^{\ell,k_{\ell}}_{n,i,j,r^*} = 1,\qquad
W^{\ell,k_{\ell}}_{n,i,j,r} = 0, ~~\forall r = 1,\dots, p_x^{\ell}p_y^{\ell}, r\neq r^*.
\end{eqnarray}
For the blocks with all zero elements, we use this $W^{\ell,k_{\ell}}_{n,i,j,r^*}=1$ to denote the ${r^*}$-th element being pooled to the upper layer. 
Therefore,
\begin{equation}
{\textstyle X^{\ell+1, k_{\ell}}_{n,i,j} = W^{\ell,k_{\ell}}_{n,i,j,r^*}  S^{\ell,k_{\ell}}_{n,i,j,r^*}} .
\end{equation}

Note that in our model, we don't explicitly have the variable $\Wmat$, and we impose in each block
\begin{eqnarray}
S^{\ell,k_{\ell}}_{n,i,j,r^*} = X^{\ell+1,k_{\ell}}_{n,i,j}, \qquad
S^{\ell,k_{\ell}}_{n,i,j,r} = 0, ~~\forall r = 1,\dots, p_x^{\ell}p_y^{\ell}, r\neq r^*.
\end{eqnarray}
We will optimize this $r^*$ for each pooling block during the model leaning process.	
The cost function for $(\ell+1)^{th}$ layer can be formulated as
{\scriptsize 
	\begin{align}
	{\cal C}^{\ell+1} &= \frac{1}{2} \sum_{n}\|\Xmat^{\ell+1}_n - \sum_{k_{\ell+1}=1}^{K_{\ell+1}} \Dmat^{\ell+1}_{k_{\ell+1}} * \Smat^{\ell+1}_{k_{\ell+1},n}\|_2^2 +\lambda \sum_{n}\|\Smat^{\ell+1}_n\|_1 \nonumber\\
	&= \frac{1}{2} \sum_{n}\|\Wmat^{\ell}_{n} \Smat^{\ell}_{n} - \sum_{k_{\ell+1}=1}^{K_{\ell+1}} \Dmat^{\ell+1}_{k_{\ell+1}} * \Smat^{\ell+1}_{k_{\ell+1},n}\|_2^2 +\lambda \sum_{n}\|\Smat^{\ell+1}_n\|_1,\nonumber
	\end{align}}
where $\Wmat^{\ell}_{n}$ (used as a function here) decides whether the element in $\Smat^{\ell}_{n}$ is pooled or not.
We further define $\tilde{\Wmat}$ being the inverse process of $\Wmat$, {\em i.e.}, determining which position of the elements in each block that $\Xmat^{\ell+1}_n $ mapped (unpooled) to.
Following this, the cost function is 
\begin{align} \label{eq:Opt_layer_W}
{\cal C}^{\ell+1} &= \frac{1}{2} \sum_{n}\left\|\Smat^{\ell}_n - \tilde{\Wmat}^{\ell}_{n}\left(\sum_{k_{\ell+1}=1}^{K_{\ell+1}} \Dmat^{\ell+1}_{k_{\ell+1}} * \Smat^{\ell+1}_{k_{\ell+1},n}\right)\right\|_2^2  {\textstyle +\lambda \sum_{n}\|\Smat^{\ell+1}_n\|_1}.
\end{align}
\paragraph{Determine the Unpooling Function $\tilde{\Wmat}$}
The above optimization problem (\ref{eq:Opt_layer_W}), if $\tilde{\Wmat}$ is given, can be solved by the CFA described in Algorithm~\ref{algo:ADMM-CFA}. 
We have developed two approaches to estimate this $\tilde{\Wmat}$.
$i$) Learn this $\tilde{\Wmat}$ in the joint training process via greedy methods. We search every candidate and select the best one (with minimum error).
$ii$) Pretrain the model without the constrain that each block has at most one nonzero element and store the indices of elements with the maximum absolute value in each block. These indices will be used as $\tilde{\Wmat}$ during the optimization of other parameters in CFA.
We have tested both and found that the second approach is more efficient in the experiments.
\paragraph{An Example of A Two-Layer Model}
For a two-layer model, we can separately optimize each layer or jointly optimize all parameters after we get the unpooling function $\tilde{\Wmat}$. Recall that the first layer is a CS model,
{\scriptsize \begin{align}
	{\cal C}^{(1)} &= \frac{1}{2} \sum_{n=1} \left\|\yv_n -\Amat \sum_{k_1=1}^{K_1} \Tmat^{(1)}_{k_1} \sv^{(1)}_{k_1,n}\right\|_2^2 +\lambda^{(1)} \sum_{n,k_1} \|\sv^{(1)}_{k_1,n}\|_1,\nonumber\\
	{\cal C}^{(2)} &= \frac{1}{2} \sum_{n}\left\|\sv^{^{(1)}}_n - \tilde{\Wmat}^{(1)}_{n}\left(\sum_{k_{2}=1}^{K_{2}} \Tmat^{(2)}_{k_{2}} \sv^{(2)}_{k_{2},n}\right)\right\|_2^2 +\lambda^{(2)} \sum_{k_2,n}\|\sv^{(2)}_{k_2,n}\|_1,\nonumber
	\end{align}}
where $\tilde{\Wmat}^{(1)}_{n}$ is obtained by pre-training, $ \Tmat^{(\ell)}$ is the matrix formulation of $\Dmat^{(\ell)}$ used in the convolution defined in (\ref{eq:Tsv}), and $\sv^{(\ell)}$ is the vectorized formulation of $\Smat^{(\ell)}$ used in the convolution.  
\paragraph{Project Upper Layer Dictionary to the Data Layer}
In the conventional deconvolutional networks~\cite{Zeiler10CVPR,Chen13deepCFA}, one needs to perform multilayer deconvolution to obtain the top layer features, which is very time consuming. Hereby, we project the upper layer dictionary element down to the bottom layer (the desired signal layer, touching $\xv_n$). One challenge is how to select the pooling map, since each dictionary will have multiple ($p_x^{\ell}\times p_y^{\ell}$) pooling maps. In order to overcome this problem, we count each candidate and select the one with maximum activations. In this case, each upper dictionary element will only have one pooling map. This same deterministic unpooling is performed for each
pooling region at a given layer, corresponding to the most probable
such map from the training data. 
Note that we already pre-store these pooling maps when we determine the unpooling function $\tilde{\Wmat}$.
Via doing this layer by layer, we can project dictionaries at each layer to the signal layer (see Figure~\ref{fig:MNIST_dict} for examples). Following this approximation, we can represent all dictionaries in the signal layer and only a single layer deconvolution is required for image reconstruction and feature extraction.

\begin{figure}[htbp!]
	\centering
	%\vspace{-1mm}
	\includegraphics[width =.8\textwidth]{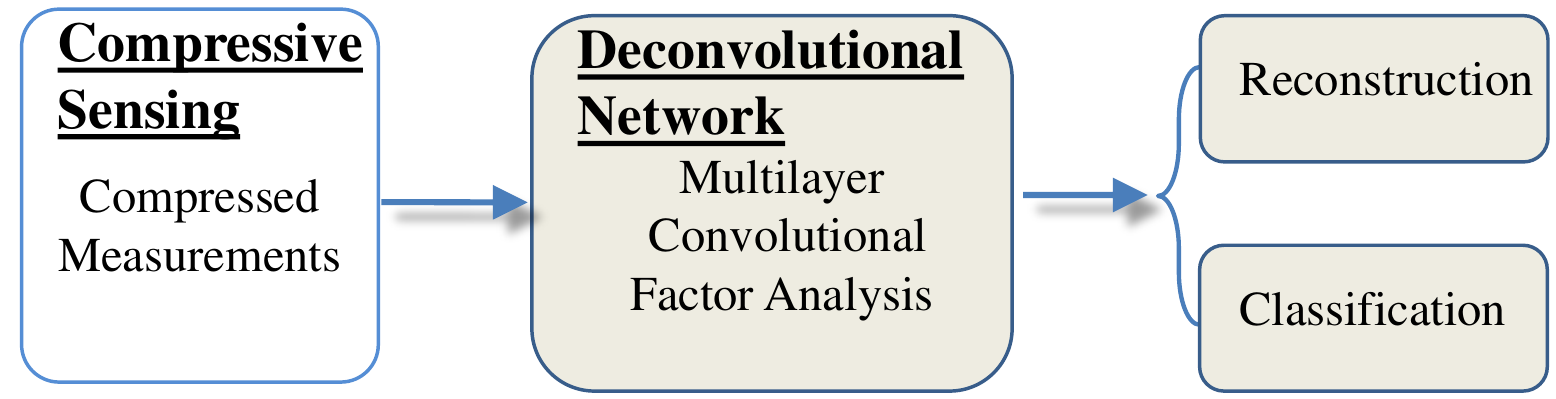}
	\vspace{-3mm}
	\caption{{\small Flow chart of our work.}}
	\vspace{-9mm}
	\label{fig:graph}
\end{figure}
\section{Joint Reconstruction and Classification \label{Sec:Joint_model}}
The deep convolutional and deconvolutional networks have achieved excellent classification results on many datasets. These networks are used to extract features $\{\Smat^{(\ell)}\}_{\ell=1}^L$ and then these features are used for classification.
In our work, we have developed a deep CS-CFA model and these features are already obtained during reconstruction. Indeed, after we get these futures, we reconstruct images as
\begin{equation}
{\textstyle {\hat{\Xmat}_n = \sum_{k_1 =1}^K \Dmat_{k_1}^{(1)} * \Smat_{k_1, n}^{(1)}}.}
\end{equation}
When performing classification, we can either superimpose an $(L+1)^{th}$ layer on the top of the deep CS-CFA model or use a separate classifier, {\em e.g.}, employing a support vector machine (SVM) on features trained by our model during reconstruction. Figure~\ref{fig:graph} depicts our joint model.

\paragraph{Joint Modeling Using Softmax}
\label{Sec:softmax}
By concatenating all features at $L^{th}$ layer, we have vectorized features $\sv_n^{(L)} \in{\mathbb R}^{N_s}$. Now we put $(L+1)^{th}$ layer on the top of the network. 
For joint classification and reconstruction task, we have labeled (compressed) data $\{\yv_n, c_n\}_{n=1}^N$, where $c_n = \{1, \dots, C\}$ considering $C$ classes in total.
Introducing the classifier weight matrix $\Hmat \in {\mathbb R}^{C \times N_s}$ and the bias vector $\alphav \in {\mathbb R}^{N_s}$, we have
\begin{align}
p(c_n = i| \sv_n, \Hmat, \alphav) = {\rm softmax}_i (\hv_i \sv_n^{(L)} + \alphav) =  \frac{\exp(\hv_i \sv_n^{(L)} + \alpha_i)}{\sum_j \exp(\hv_j \sv_n^{(L)} + \alpha_j)},
\end{align}
where $\hv_i$ denotes the $i^{th}$ row of the weight matrix $\Hmat$ and $\alpha_i$ symbolizes the $i^{th}$ element of the vector $\alphav$.
These weights $\{\Hmat, \alphav\}$ can be learned jointly with the CFA network, thus constituting a supervised CS-CFA model.
Similarly, a $C$-class SVM can also be used~\cite{yang09CVPR}.

\section{Experimental Results \label{Sec:result}}
We conduct our proposed algorithm on digit and face datasets. Firstly, we verify the CS inversion performance of our model on a small subset of MNIST.
Then we conduct our model for joint classification and reconstruction on the complete MNIST dataset.

The algorithm is implemented in MATLAB. The model is randomly initialized and we run 200 iterations or terminate when the relative measurement error is below $10^{-5}$, which one comes first.
Regarding the parameter setting of $\{\beta,\eta\}$, we have found that $\beta$ can be set to a constant, {\em e.g.}, $10^{-7}$ used in MNIST. $\eta$ should be firstly set to a small number ({\em e.g.}, $10^{-3}$) and then increases in each iteration and finally set to a maximum number ({\em e.g.}, 5).
During the training process, this increase of $\eta$ will be an inter loop beside the outer loop as demonstrated in Algorithm~\ref{algo:CS_CFA}.
We further observe that in each layer, each slice of $\Dmat$ can be updated simultaneously, while $\Smat$ should be updated slice by slice, where each slice corresponds to one slice of the dictionary in that layer.
\subsection{CS Reconstruction}
\paragraph{MNIST}
The MNIST data (\url{http://yann.lecun.com/exdb/
	mnist/}) has 60,000 training and 10,000
testing images, each $28\times28$, for digits 0 through 9.
We randomly select 100 digits (10 for each) for CS reconstruction.
The sensor matrix is of dimension $\Amat \in {\mathbb R}^{M \times 
	(N_x N_y)}$	and we define
\begin{equation}
{\rm CSr} = \frac{M}{N_x N_y} = \frac{\text{number of row in } \Amat}{\text{number of column in } \Amat}.
\end{equation}
Each column of the sensing matrix is drawn from Gaussian and normalized~\cite{He09SPT}.
We conduct the experiments by both training the convolutional dictionary $\Dmat$ from another set of digits, and learning $\Dmat$ from the measurements directly, thus {\em in situ}. The dictionary size is set to $7\times 7 \times 16$ and a single layer model is used.

\begin{wrapfigure}{r}{0.32\textwidth}
	\vspace{-3mm}
	\centering
	\includegraphics[scale=0.29]{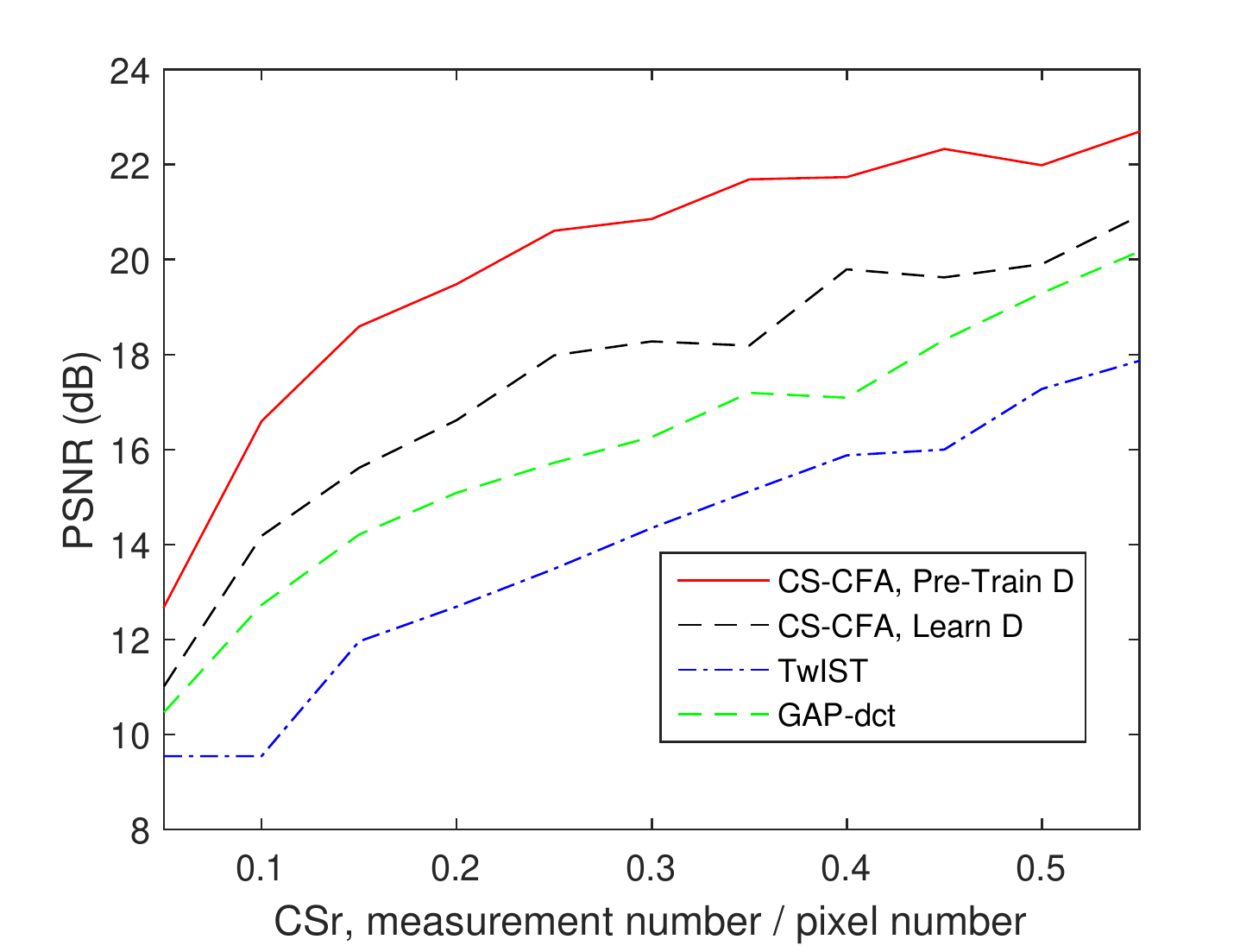}
	\vspace{-3mm}
	\caption{{\small Average PSNR of reconstructed results  versus CSr with different algorithms on MNIST dataset. }}
	\label{fig:MNIST_rec_psnr}
	\vspace{-5mm}	
\end{wrapfigure}
\begin{figure}[tbp!]
	\centering
	\vspace{-3mm}
	\includegraphics[width = .75\textwidth]{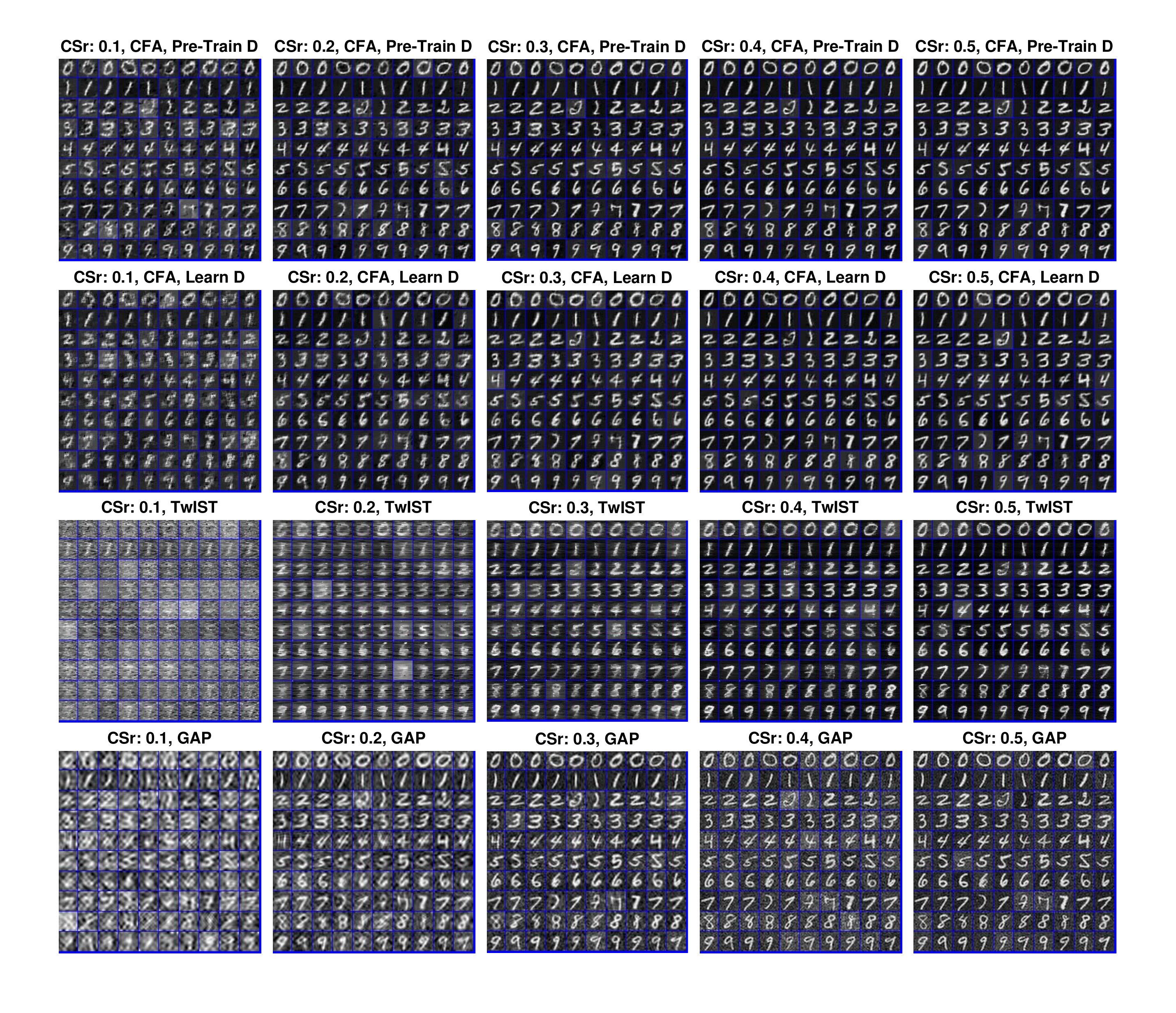}\\
	\vspace{-4mm}
	\caption{{\small Reconstruction results of MNIST dataset at various CSr. Row 1: proposed CS-CFA with pre-trained dictionary; row 2: proposed CS-CFA with {\em in situ} learned dictionary (from the measurements directly); row 3: TwIST; row 4: GAP.}}
	\vspace{-9mm}
	\label{fig:MNIST_rec}
\end{figure}

The average PSNR of the reconstructed digits versus CSr is plotted in Figure~\ref{fig:MNIST_rec_psnr}. The reconstructed digits compared with other algorithms are demonstrated in Figure~\ref{fig:MNIST_rec}. 
TwIST~\cite{Bioucas-Dias2007TwIST} with total variation (TV) regularization, and GAP~\cite{Liao14GAP,Yuan16ICIP_GAP} (imposing sparsity on DCT coefficients) are used as baselines.
It can be observed that our proposed algorithm, both learned dictionary {\em in situ} and with pre-learned dictionary, performs better than other algorithms. If training data are available, the algorithm performs best, $i.e.$, a 2dB increase compared to {\em in situ} learned dictionary.
TwIST performs worst (we also tried TVAL3~\cite{Li13COA}, which performs even worse than TwIST). One proper reason is that these digital images are small and TV may not be the best regularizer. GAP with DCT performs better than TwIST. However, it is not as good as the proposed CS-CFA. 
Similar results can be found in wavelet based algorithms. Again, this may due to the image size and this phenomenon has also been observed by~\cite{Chen10SPT} when they compared their algorithm with the Bayesian algorithms~\cite{He10SPL}. 
%
%\begin{figure}[htbp!]
%	\centering
%	\vspace{-4mm}
%	\includegraphics[width =.38\textwidth]{Figures/psnr_mnist.pdf}
%	\vspace{-3mm}
%	\caption{Average PSNR of reconstructed results  versus CSr with different algorithms on MNIST dataset. }
%	\vspace{-4mm}
%	\label{fig:MNIST_rec_psnr}
%\end{figure}
%
\begin{wrapfigure}{r}{0.45\textwidth}
	\vspace{-1mm}
	\centering
	\includegraphics[width = .2\textwidth]{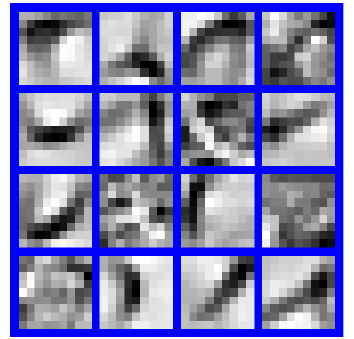}~
	\includegraphics[width = .2\textwidth]{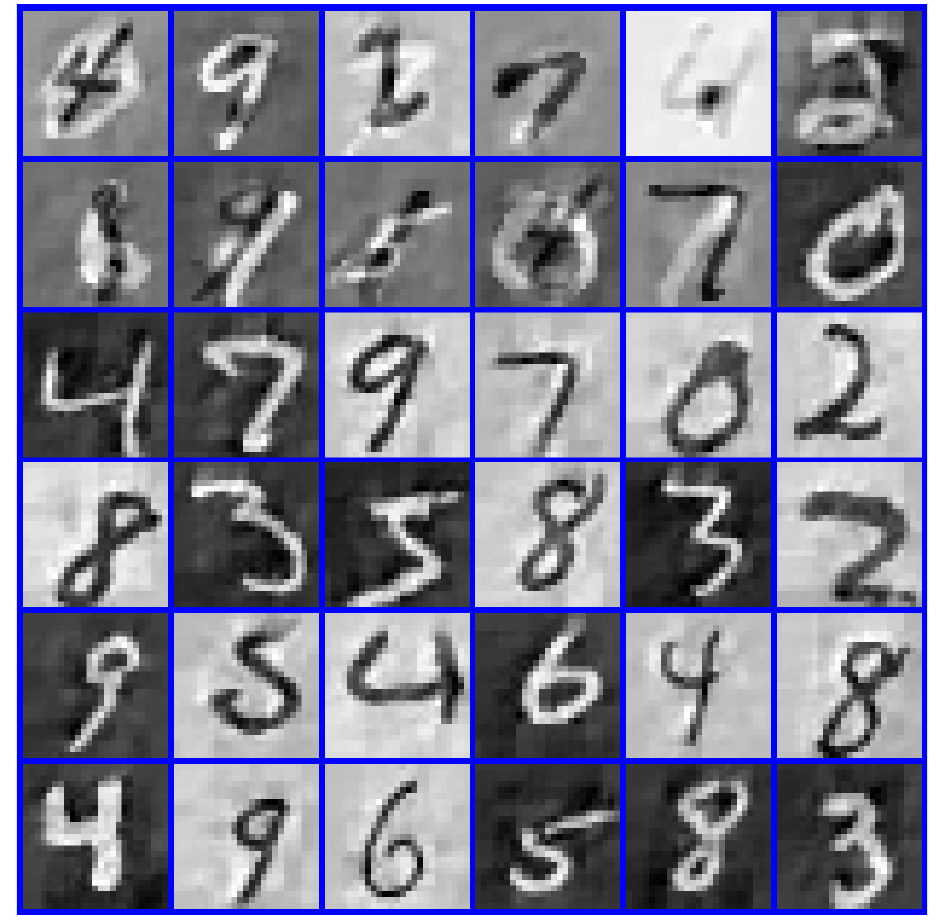}
	\vspace{-3mm}
	\caption{{\small Selected dictionary elements trained on MNIST. Left: layer 1, right:layer 2.}}
	\label{fig:MNIST_dict}
	\vspace{-7mm}	
\end{wrapfigure}
\paragraph{Multilayer Model for Low CSr}
Now we show results of our deep model. Specifically, we employ a two-layer model on MNIST. For the first layer, we use the dictionary size of $9\times 9 \times 16$ and the pooling block is of size $3\times 3$, and the dictionary at the second layer is of size $7\times 7\times 64$.
The learned dictionary, visualized in the data layer is shown in Figure~\ref{fig:MNIST_dict}.
It can be seen that the bottom layer dictionary looks like edges and the layer 2 dictionary looks like digits when projected to the signal layer.

\begin{wrapfigure}{r}{0.5\textwidth}
	\vspace{-5mm}
	\centering
	\includegraphics[width = .5\textwidth]{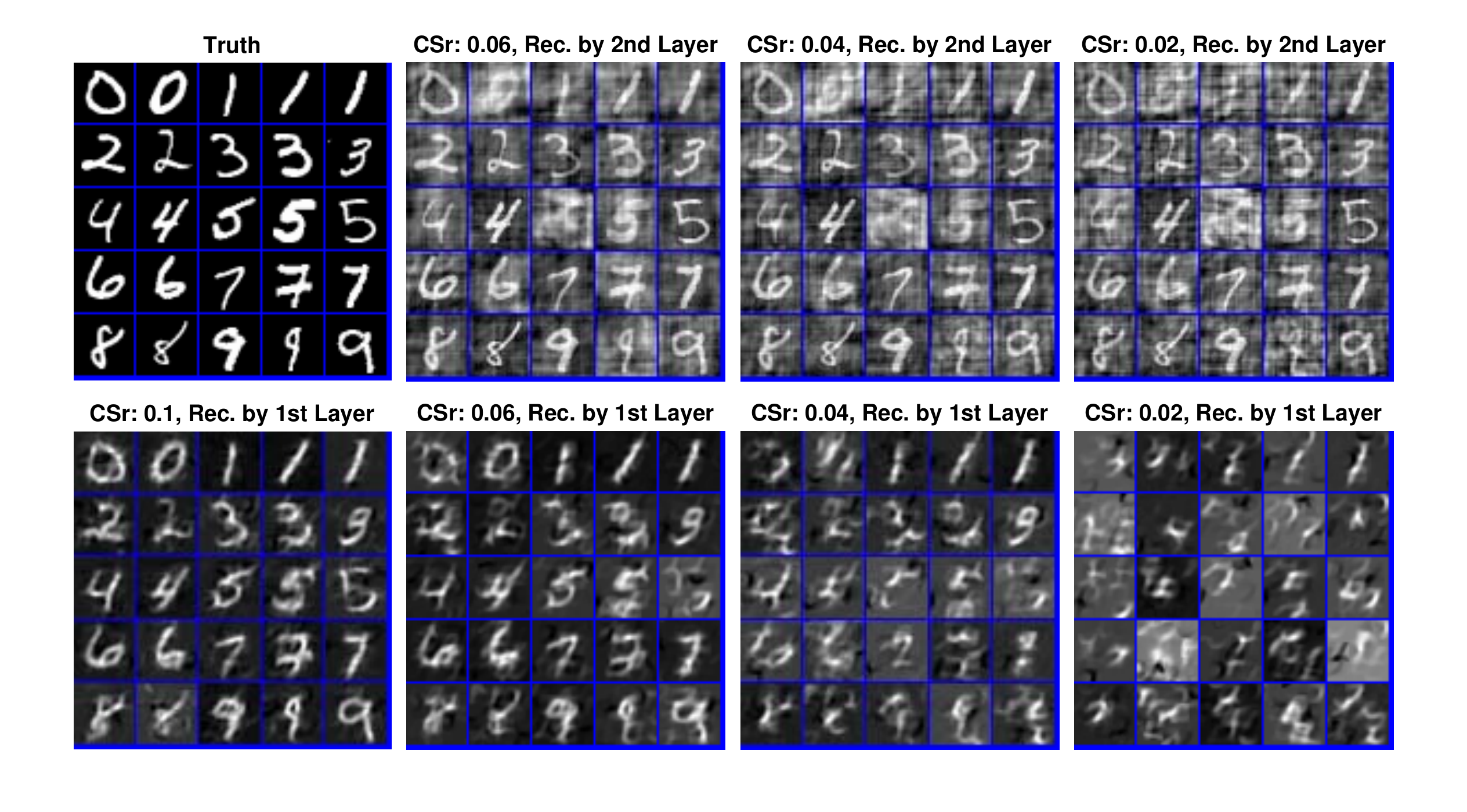}
	\vspace{-3mm}
	\caption{\small{Reconstruction results using first layer (row 2) and second layer (row 1) dictionaries at low CSr $\in [0.02,0.06]$.}}
	\label{fig:MNIST_rec_2nd}
	\vspace{-7mm}	
\end{wrapfigure}
We observe that when the CSr is relatively large ($>0.1$), the first layer model always provides better results than the deep model since the error will be accumulated when the model goes deep. 
However, when the CSr is extremely small, a deep model can perform better. 
We show 25 reconstructed digits at CSr $\in[0.02, 0.1]$ in Figure~\ref{fig:MNIST_rec_2nd} as an example.
It can be observed that though the background of the reconstruction using second layer dictionary looks noisy, we can identify the outlines of digits. By contrast, the reconstruction results using first layer dictionary have random noise everywhere, thus hard to recognize. 
%
%\begin{figure}[htbp!]
%	\centering
%	\vspace{-3mm}
%	\includegraphics[width = .5\textwidth]{Figures/mnist_lowCSr.pdf}
%	\vspace{-7mm}
%	\caption{Reconstruction results using first layer (row 2) and second layer (row 1) dictionaries at low CSr $\in [0.02,0.06]$.}
%	\vspace{-6mm}
%	\label{fig:MNIST_rec_2nd}
%\end{figure}
%
\begin{wrapfigure}{r}{0.5\textwidth}
	\vspace{-5mm}
	\centering
	\includegraphics[width =.5\textwidth]{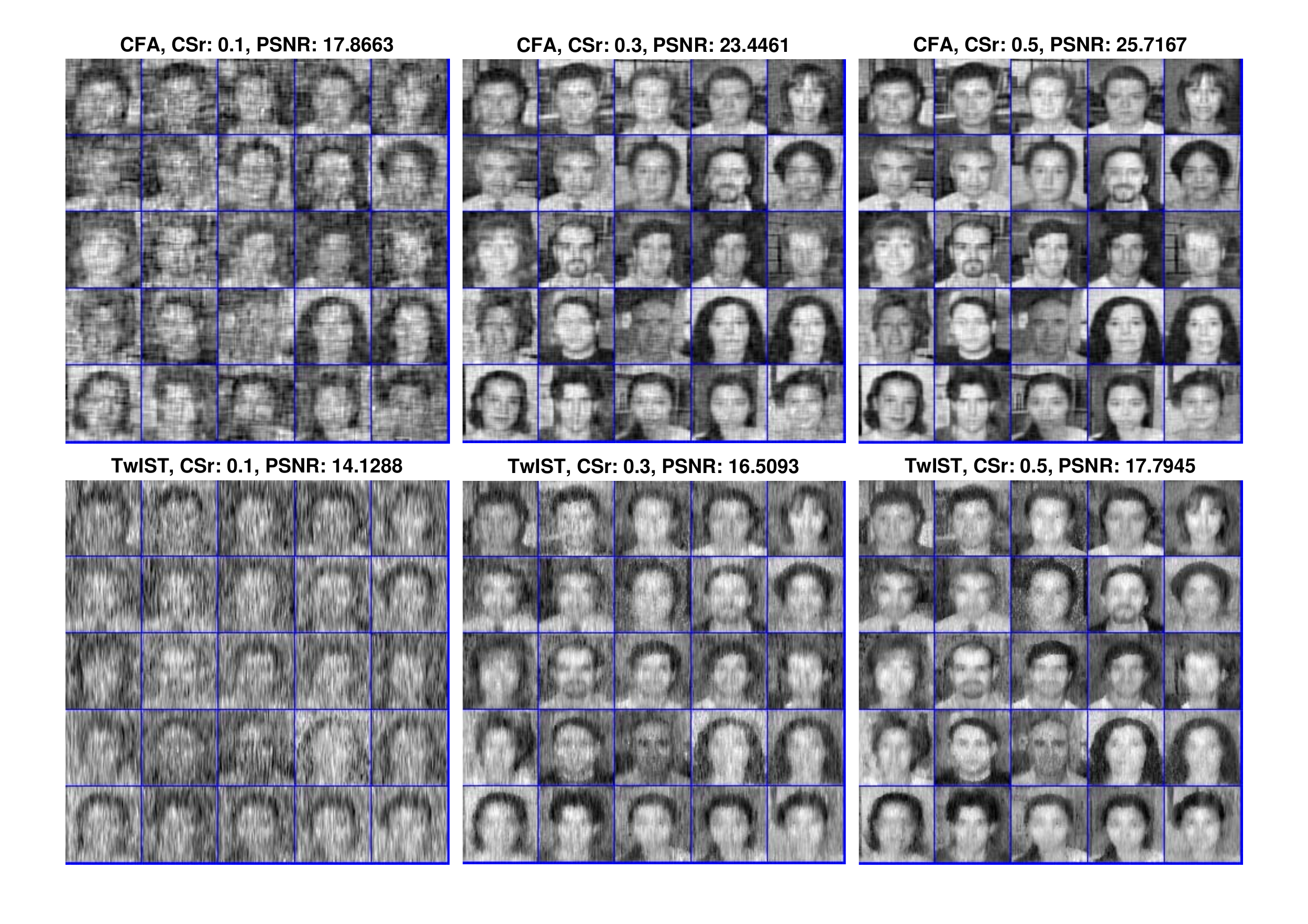}
	\vspace{-3mm}
	\caption{\small{Selected reconstructed images and average PSNR (dB) of ``face$\_$easy" category in Caltech101 data at various CSr. Row 1: proposed CS-CFA, row 2: TwIST.}}
	\label{fig:face_rec}
	\vspace{-7mm}	
\end{wrapfigure}
\paragraph{Face data}
In addition to the digital dataset, we also test our algorithm on the ``face$\_$easy"
category in Caltech256~\cite{caltech256}. There are 435 images and we first convert them to grayscale and then resize all of them to $64\times 64$. 
In this dataset, we use the permuted Hadamard matrix as the sensing matrix~\cite{Li13COA,Huang13ICIP} in order to speed up the inference.
The dictionary size is set to $13\times 13 \times 16$ in a single layer model. We use half of the images to train our CFA model and test on the rest. 
In Figure~\ref{fig:face_rec}, we show 25 reconstructed images compared with TwIST. Again, our model performs better.
Though not reported here, similar to digits, when going deep, using upper layer dictionary can provide better reconstruction results at extremely low CSr. However, when the CSr is not that small, reconstruction results using the first layer dictionary are the best. 
One reason is that only the first layer dictionary includes details of the images, while the upper layer dictionary usually captures the structure.

%
%\begin{figure}[htbp!]
%	\centering
%	\vspace{-1mm}
%	\includegraphics[width =.5\textwidth]{Figures/face_25_2.pdf}
%	\vspace{-6mm}
%	\caption{Selected reconstructed images and average PSNR (dB) of ``face$\_$easy" category in Caltech101 data at various CSr. Row 1: proposed CS-CFA, row 2: TwIST.}
%	\label{fig:face_rec}
%\end{figure}

\begin{table}[htbp!]
	\vspace{-2mm}
	\caption{Reconstruction PSNR (dB) and classification accuracy (\%) at various CSr on MNIST.}
	\centering
	\begin{small}
		\begin{tabular}{c|ccccccccc|c}		
			CSr & 0.1  & 0.2 & 0.3 & 0.4  & 0.5 & 0.6 & 0.7 & 0.8 & 0.9 & no CS\\
			\hline 
			Reconstruction PSNR & 15.65 & 19.73 & 21.55 & 22.51 & 23.53 & 24.02 & 24.22 & 24.63 & 24.84 & -\\
			\hline
			Softmax & 75.88 & 90.88& 92.89 &93.78 & 94.47 & 94.46 & 94.35 & 94.38 & 94.39 & 93.84\\
			Linear SVM & 70.30 & 88.57& 91.64 &93.45 & 93.79 & 93.76 & 93.12 & 93.52 & 93.56 & 92.13\\
			Nonlinear SVM & 77.62 & 92.38& 94.93 &95.29 & 96.37 & 96.77 & 96.84 & 96.82 & 96.79 & 96.32\\
			\hline 
		\end{tabular}
	\end{small}
	\vspace{-2mm}
	\label{Table:sim_PSNR}
\end{table}
\subsection{Reconstruction and Classification}
Now we conduct our model on the complete MNIST dataset for joint reconstruction and classification. We perform our CFA model in Algorithm~\ref{algo:ADMM-CFA} on the 60000 training digits to extract features and these features are sent to classifiers for training.  
During testing, the features are extracted directly from the compressed measurements of 10000 digits.
The reconstruction and classification results are summarized in Table~\ref{Table:sim_PSNR}. It can be observed that our model can simultaneously reconstruct the images and classify the digits using {\em compressed measurements}.
The softmax classifier introduced in Section~\ref{Sec:softmax} is utilized.
Regarding the classification performance of our model, according to the UFLDL tutorial\footnote{ \url{http://ufldl.stanford.edu/wiki/index.php/Exercise:Softmax_Regression}}, the classification accuracy of the original data using softmax is $92.6\%$, while we have achieved $92.89\%$ when CSr = $0.3$. For comparison, we also present the classification results of both linear and nonlinear SVM~\cite{CC01a}. Without comparing these different classifiers, we observe that for every classier, we can achieve comparable classification result to the original data with $30\sim40\%$ compressed measurements, which reduced more than half of the data volume.
We further notice that due to the randomness introduced by the compressive sensing matrix, the classification results are improved compared to the original data. Specifically, the softmax classification accuracy is $93.84\%$ using CFA features extracted from the original data, but we get $>94\%$ accuracies using CFA features extracted from the {\em compressed} data when CSr$>0.4$. Similar observations can be found in other classifiers. This is in agreement with the  theory recently developed in~\cite{Huang15ICASSP,Huang15NIPS}.

It is worth noting that we are not aiming to compete with any other deep models~\cite{Zeiler14ECCV,donahue2013decaf,GoogleNet,HeSSPnet}, that dedicate to get high classification performance.
% though our multilayer model can achieve better results than those reported in Table~\ref{Table:sim_PSNR}.
Our target is to show that using {\em compressed data}, which reduced data volume significantly, can still achieve comparable performance to the original data, which are the input of other deep models. We believe that by using a very deep model in our regime, we can get comparable performance to other models.
However, the reconstruction results of the deep model may be not as good as those reported here, while this is the pivotal task in CS.

\section{Conclusion and Discussion}
We have developed a novel {\em compressive sensing} convolutional factor analysis model based on the alternating direction method of multipliers paradigm. The model learns features as well as dictionaries from the compressed measurements. 
Joint reconstruction and recognition is straightforward since the features are used to reconstruct the desired signals. 
This bridges the gap between compressive sensing and deep deconvolutional learning.
A stochastic unpooling process is employed to link adjacent layers when a deep generative model is constructed.
Via this stochastic unpooling and the approximate process of projecting upper dictionary down to the data layer, only a single layer deconvolution is required for reconstruction and testing.
The proposed model provides excellent results on the compressive sensing inversion task through {\em in situ} learn dictionary for joint recovery of multiple signals.
% ,  are required for joint learning and recovery.
%Otherwise, we can pre-train the dictionary using related datasets. 
We have demonstrated that for both image reconstruction and classification tasks, the proposed compressive convolutional factor analysis algorithm can dramatically ($>60\%$) reduce the data volume.

Regarding the computational time, deconvolution during inversion takes longer time than the wavelet, DCT or total variation based algorithms~\cite{Bioucas-Dias2007TwIST,Li13COA,Liao14GAP}.
However, it is comparable to the advanced patch-based algorithms~\cite{Dong14TIP,Huang14TIP,Mertzler14Denoising}, which usually need the images to be large in order to extract sufficient patches.
Our algorithm is not only an alternative inversion approach for compressive sensing,  but also provides excellent reconstruction and classification results with very limited measurements.
Our future work will test the proposed model on real categorical data (both images and videos) captured by compressive sensing cameras~\cite{Yuan16AO,Yuan16SJ,Yuan16BOE,Yuan15Lensless,Yuan_16_OE,Sun16OE,Yuan&Pang16_ICIP,Yuan15JSTSP,Yuan15Lensless}.

% \section*{Acknowledgements}
%
% \textbf{Do not} include acknowledgements in the initial version of
% the paper submitted for blind review.
%
% If a paper is accepted, the final camera-ready version can (and
% probably should) include acknowledgements. In this case, please
% place such acknowledgements in an unnumbered section at the
% end of the paper. Typically, this will include thanks to reviewers
% who gave useful comments, to colleagues who contributed to the ideas,
% and to funding agencies and corporate sponsors that provided financial
%\clearpage
%
\bibliographystyle{acm}

%\bibliography{hmc,reference_sideinfor,reference_ECCV,aistats2016}
%
%
\end{document}